%% file: main.tex
\documentclass[runningheads]{llncs}
\usepackage{graphicx}
\input{preamble}

\begin{document}
\title{Learning from Partial Label Proportions \\
for Whole Slide Image Segmentation}
%
%
\author{Shinnosuke Matsuo\inst{1} \and
Daiki Suehiro\inst{1} \and
Seiichi Uchida\inst{1} \and
Hiroaki Ito\inst{2} \and \\
Kazuhiro Terada\inst{2} \and
Akihiko Yoshizawa\inst{2} \and
Ryoma Bise\inst{1}
}

\authorrunning{S. Matsuo et al.}

\institute{Kyushu University, Fukuoka, Japan \and
Kyoto University, Kyoto, Japan}
\maketitle              
\begin{abstract}
In this paper, we address the segmentation of tumor subtypes in whole slide images (WSI) by utilizing incomplete label proportions. Specifically, we utilize `partial' label proportions, which give the proportions among tumor subtypes but do not give the proportion between tumor and non-tumor. Partial label proportions are recorded 
as the standard diagnostic information by pathologists, and we, therefore, want to use 
them for realizing the segmentation model that can classify each WSI patch into one of the tumor subtypes or non-tumor. We call this problem ``learning from partial label proportions (LPLP)'' and formulate the problem as a weakly supervised learning problem. Then, we propose an efficient algorithm for this challenging problem by decomposing it into two weakly supervised learning subproblems: multiple instance learning (MIL) and learning from label proportions (LLP). These subproblems are optimized efficiently in an end-to-end manner. The effectiveness of our algorithm is demonstrated through experiments conducted on two WSI datasets.

\end{abstract}
%
%

\section{Introduction\label{sec:intro}}

\begin{figure}[t]
\centering
 \includegraphics[width=1.\linewidth]{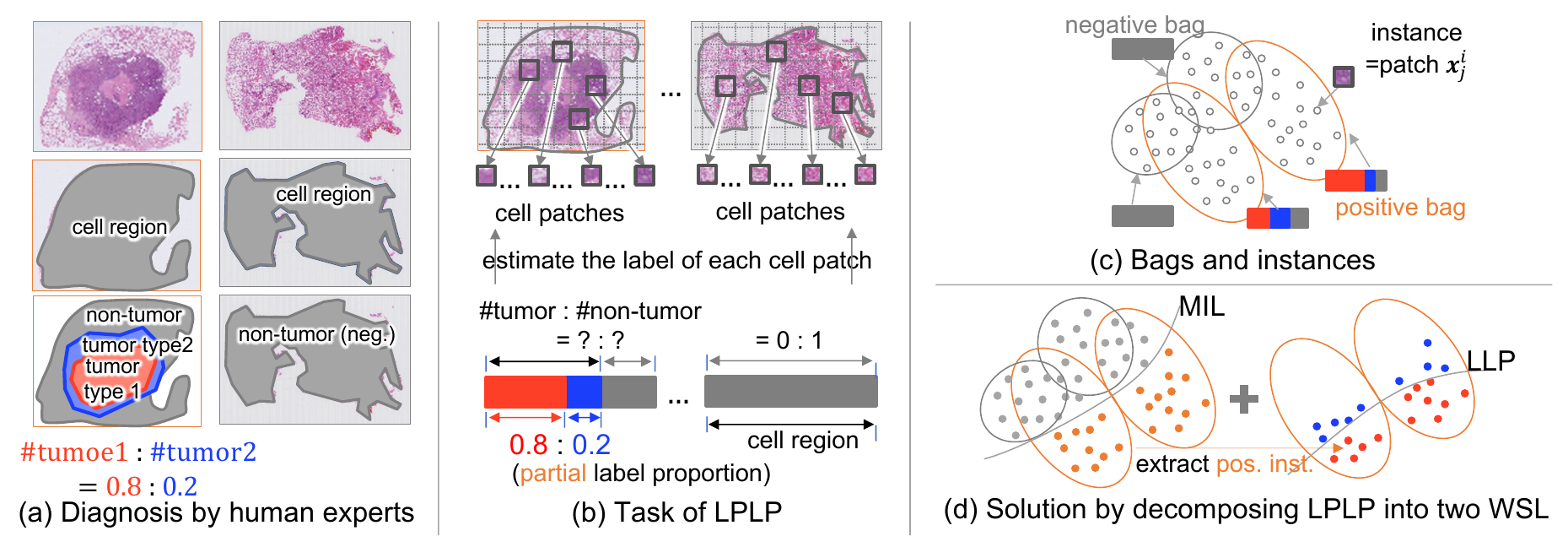}
\caption{Overview of our work.}
\label{fig:problem_setting}
\end{figure}

Automated segmentation of whole slide images (WSI) by tumor subtypes is crucial in aiding pathologists with accurate tumor diagnosis. Since WSI often has an extremely high resolution, its segmentation task can be treated as a patch-wise classification task. In other words, our goal is to classify each patch from the cell region into one of the $C$ tumor subtype classes or the non-tumor class.
\par
A unique situation of automated WSI segmentation is that we can easily collect label proportions among $C$ tumor subtype classes for each WSI in many pathological diagnoses.
As shown in Fig.~\ref{fig:problem_setting}(a), when $C=2$, pathologists' interest in their diagnosis is determining the proportion among two tumor subtypes, such as $0.8:0.2$. 
For instance, in clinical practices like chemotherapy~\cite{travis2020iaslc}, the PDL1 positive rate~\cite{Liu2021,roach2016development,Widmaier2020}, and the diagnosis of lung tissue after surgery~\cite{Dacic2021,moreira2020grading}, the determination of a condition's grade depends on the proportion of tumor subtypes excluding non-tumor areas. 
\par

For realizing patch-wise classification, however, the proportions only among tumor classes are ambiguous information because they do not care about the non-tumor region. 
In other words, we can only know the {\em partial} label proportion.
Assume $(0,8, 0.2)$, where $0.8$ and $0.2$ show the proportion between the two tumor subtypes. Note that the proportion of the non-tumor region is unknown.  
Then, if the proportion between tumor and non-tumor is found to be $0.5:0.5$, the actual label proportion is $(0,4, 0.1, 0.5)$. 
\par

This work tackles a new machine-learning problem, {\em learning from partial label proportions} (LPLP). As shown in Fig.~\ref{fig:problem_setting}(b), by using partial label proportions
of WSI images, we train a cell patch classifier to classify a patch into one of $C+1$ classes. This problem is not solvable without any hint of non-tumor patches; this is simply because the partial label proportions do not give any information about non-tumors. Fortunately, in practical WSI diagnosis scenarios, pathologists often encounter 
``totally healthy'' WSIs, like the right WSI in Fig.~\ref{fig:problem_setting}(a). Since this WSI contains no tumor cells, all its cell patches are guaranteed as non-tumor patches.
As detailed in the following, we will fully utilize such WSIs to make LPLP a solvable problem.
\par

As shown in Fig.~\ref{fig:problem_setting}(c), LPLP is formulated by instances, bags, and partial label proportions. An instance corresponds to each cell patch and thus belongs to one of the $C+1$ classes. 
Each bag can be positive or negative; a {\em positive bag} contains instances of tumor patches and, in general, also contains instances of non-tumor patches. In contrast, 
a {\em negative bag} contains only instances of non-tumor. A healthy WSI forms a negative bag. A bag is associated with a partial label proportion.
For example, if $C=2$ and the tumor region contains $80\%$ of tumor type1 (red) and $20\%$ of type2 (blue), then the partial label proportion of tumor subtype classes is given as $(0.8, 0.2)$. In contrast, we do not know the proportion of the non-tumor class except when the bag is negative.
\par

The highlight of this paper is that we decompose LPLP into two weakly supervised learning (WSL) subproblems: multiple instance learning (MIL) and learning from label proportions (LLP) as shown in Fig.~\ref{fig:problem_setting}(d).
MIL is known as a WSL problem using binary bag labels (positive or negative). In our MIL as the subproblem, we aim to classify the instances for positive (tumor) or negative (non-tumor) as shown in Fig.~\ref{fig:problem_setting}(d) left.
LLP is a WSL problem using label proportions for {\em all} classes. To complement the lack of the proportion information of negative instances, we use the inference of the above MIL. More precisely, our LLP as subproblem is for the subbags of the positive bags, which contains only positive instances as shown in Fig.~\ref{fig:problem_setting}(d) right. Note that these two WSLs are not disjoint; we design a neural network that can simultaneously solve both MIL and LLP problems end-to-end.
\par

\section{Related work\label{sec:relatedwork}}
\begin{table*}[t]
\small
\caption{Our problem setting and conventional WSL.  \label{tab:related}}
\centering
\begin{tabular}{llll}
\hline
                 & LPLP (Ours)               & MIL                                                                   & WSL other than MIL          \\ \hline
Labels           & {\bf Partial} label proportions & Bag-level Pos/Neg label                  & Scribbles, etc.    \\ \hline
\begin{tabular}[c]{@{}l@{}}Labeling\\ cost ($\downarrow$)\end{tabular} &
  \begin{tabular}[c]{@{}l@{}}Low \\ (easily available)\end{tabular} &
  \begin{tabular}[c]{@{}l@{}}Low \\ (easily available)\end{tabular} &
  \begin{tabular}[c]{@{}l@{}}Relatively High \\ (with extra-annotation)\end{tabular} \\ \hline
Objective & Segmentation & Bag-level classification & Segmentation \\ \hline

Comparability     & \multicolumn{3}{c}{\bf No fair comparison is possible due to the above differences} \\ \hline
Existing methods & \textbf{None}                         & ~\cite{javed2022additive,shao2021transmil} etc. & ~\cite{Anklin2021LearningWS,chan2019histosegnet} etc.   \\ \hline
\end{tabular}
\end{table*}

\noindent{}
\textbf{Weakly-Supervised Learning (WSL) for Whole Slide Images (WSI). \label{related:WSL}}
WSLs for WSI are summarized in Table~\ref{tab:related}.
The most representative learning problem in WSL for WSI is multiple instance learning (MIL), where the positive (tumor) or negative (non-tumor) label is given to a bag
~\cite{HashimotoCVPR2020,javed2022additive,li2021dual,LVMICCAI2022,shao2021transmil}.
These studies' objective differs from ours; most MIL methods aim for bag-level classification but do not segment subtypes (instance-level classification).
To use an AI system in clinical, showing segmentation results is also important for explainability.
Recently, the segmentation from bag-level labels has been challenged by~\cite{qu2022bi,SILVARODRIGUEZ2022105714}; however, these methods assume the binary class 
(tumor or non-tumor)
and thus does not work for multiple classes, including tumor subtypes.
Generally, classifying subtypes is significantly more challenging than binary classification due to the similar appearances of the tumor subtypes.
\par

Several WSL approaches other than MIL have also been proposed.
For example, SEGGIN~\cite{Anklin2021LearningWS} trains a network using scribbles, which can be considered as partial instance (pixel)-level labels, and HISTOSEGNET~\cite{chan2019histosegnet} performs pixel-level WSI segmentation from patch-level labels. These methods do not use the proportions of tumor subtypes in a WSI and require extra annotations. 
In contrast, the proportions are recorded as diagnosis information as mentioned in the introduction and, thus, are easily available. 
To the authors’ best knowledge, this paper is the first work to deal with the partial label proportion.
\par

\noindent{}
\textbf{Learning from label proportions (LLP). \label{related:LLP}}
Learning from label proportions (LLP)~\cite{ardehaly2017co,asanomi2023,DulacArnoldG2020,liu2019learning,matsuo2023,okuo2023miccai,tsai2020,yang2021two,ShiY2020} is a WSL task given bags and their label proportions. For example, if a bag contains 150, 30, and 20 instances of classes 1, 2, and 3, respectively, the proportion is $(0.75, 0.15, 0.10)$. The goal of LLP is to obtain an instance classifier from bags and their label proportions. \par

The most methods of LLP are based on a proportion loss to train a deep neural network, which predicts the class of an instance.
The proportion loss is a bag-level cross-entropy between the ground truth and the label proportions estimated by averaging the probability outputs in a bag.
These methods require proportions of `all' the class.
In other words, applying LLP to this LPLP problem requires the proportion between tumor and non-tumor regions, which incurs additional labeling costs.
\color{black}

\section{Learning from partial label proportions (LPLP)}

LPLP is a novel and ``more weakly'' supervised learning problem than LLP because the label proportion of a certain class is always unknown in all bags. Hereafter, we call the unknown class ``the negative class'' and the other $C$ ``positive classes.''
Assume $n$ training (positive) bags $\mathcal{B}_{\mathrm{pos}}=\{\bB^i\}_{i=1}^n$, 
each of which contains instances of $C+1$ classes.
LPLP aims to train an instance classifier for $(C+1)$ classes
by using the partial label proportions, which are the label proportions of positive classes, $\bp^i = (p_1^i, \ldots, p_C^i) \in \Delta^C, i=1,\ldots,n$, where $\Delta^C$ is the $C$-probability simplex. 
Note that the proportion between the positive classes and the negative class is unknown. 

Solving the above LPLP problem is impossible as it is because no information is given for the negative class. 
Fortunately, negative bags $\mathcal{B}_{\mathrm{neg}}=\{\bB^i\}_{i=n+1}^m$ that {\em only} contain instances of the negative class 
are available from easily prepared healthy WSIs.
The negative bags can give a hint to discriminate the negative instances from the positive instances.
All $m$ training bags are one of these two types, i.e. $\mathcal{B} = \mathcal{B}_\mathrm{pos}$ + $\mathcal{B}_\mathrm{neg}=\{\bB^i\}_{i=1}^m$.

In summary, the problem setup of LPLP can be rewritten as follows: 
given positive training set $\{\bB^i, Y^i, \bp^i\}_{i=1}^n$ and negative training set $\{\bB^i, Y^i\}_{i=n+1}^m$, training an instance classifier for $(C+1)$ classes, composing $C$ positive classes and negative class.
Here, the bag-level binary label $Y^i \in \{0,1\}$, takes 0 if the bag $\bB^i$ is a negative bag, otherwise 1.

\section{Simultaneous learning of MIL and LLP for LPLP}

\begin{figure*}[t]
\begin{center}
   \includegraphics[width=.99\linewidth]{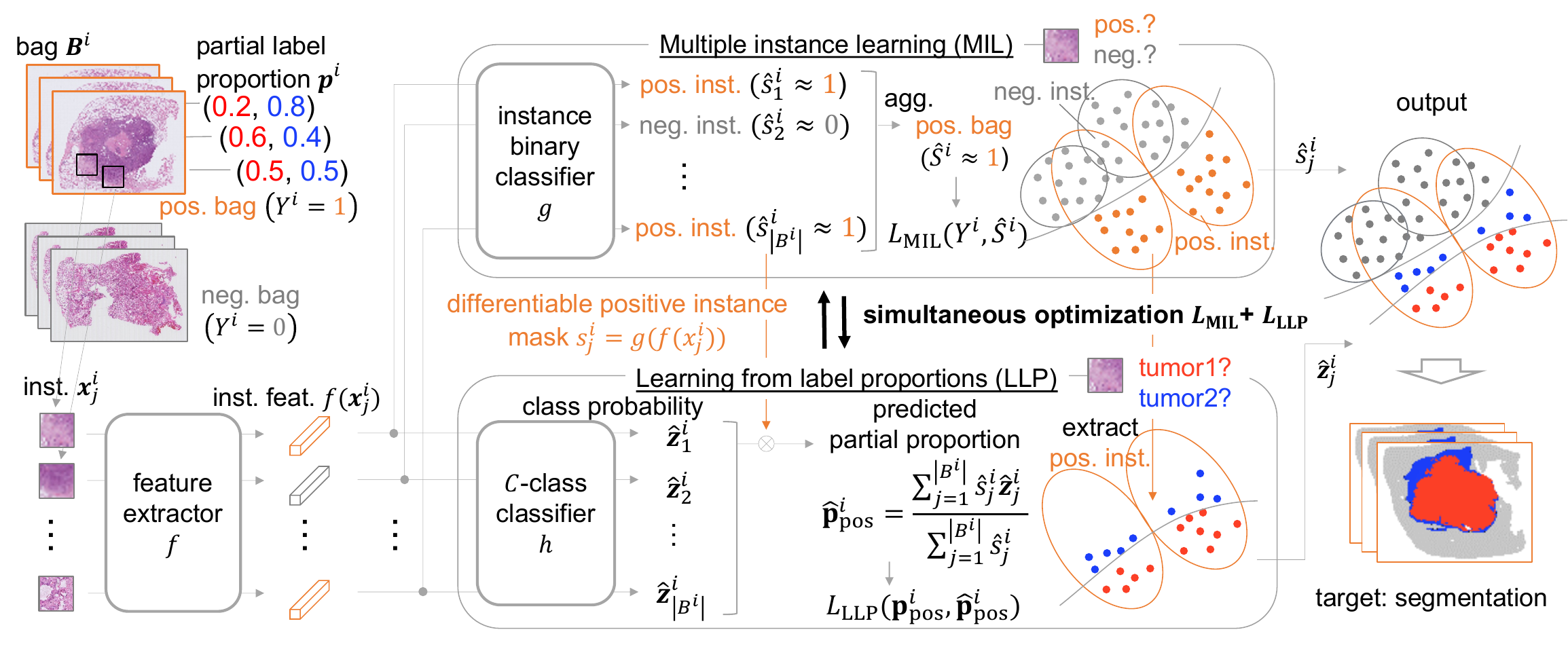}
\end{center}
\caption{Overall architecture of the proposed model. 
}
\label{fig:architecture}
\end{figure*}

Using additionally obtained the bag-level binary class information, we decompose the challenging WSL problem LPLP into two simpler WSL problems: multiple instance learning (MIL) and learning from label proportion (LLP).
As shown in Fig.~\ref{fig:architecture}, the proposed method consists of the MIL and LLP modules. In the MIL module, given bag-level binary labels, it aims to train an instance-level binary classifier $g$ that classifies an instance $x_j^i$ into positive or negative.
LLP aims to train an instance classifier $h$ for $C$ positive classes using a set of bags only containing positive instances extracted by the MIL module and their partial proportions.

If we simply apply this two-stage framework, the classification performance depends on the first step (MIL); the errors in MIL estimation might affect the second step (LLP). This is because the positive proportion is shifted if the bag contains noisy negative instances. 

To avoid this problem, we simultaneously optimize both MIL and LLP end-to-end. The MIL module classifies an instance into a positive or negative instance. Then, the positive instance is further classified into either of the $C$ positive classes while ignoring negative classes by multiplying the outputs of the LLP module by the positive instance masks. It enables to training of both modules simultaneously.


\noindent \textbf{MIL module. }
In the MIL module of our method, we utilize mi-Net (output-aggregation-MIL)~\cite{wang2018revisiting}, which estimates the bag-level label by aggregating the outputs of the feature extractor $f$ and the instance classifier $g$ for each instance in a bag.
Moreover, the classifier $g$ generates the differentiable positive instance scores (masks) $\hat{s}^i_j=g(f(x^i_j))\in[0,1]$, which complement the lack of negative proportions in positive bags for the LLP module.
The differentiable positive instance scores are close to 1 for positive instances and 0 for negative instances. All of the instance scores $\{\hat{s}^i_j\}_{j=1}^{|\bB^i|}$ in a bag are aggregated using an aggregation function, such as the mean-pooling, max-pooling, and LogSumExp (LSE)-pooling, to obtain the bag score $\hat{S}^i$. 
A bag-level binary cross-entropy is used for the loss function of the MIL task as follows:
\begin{align}
\mathcal{L}_\mathrm{MIL}(Y^i, \hat{S}^i) = -Y^i \log \hat{S}^i - (1-Y^i) \log (1-\hat{S}^i).
\end{align}

In the output-aggregation MIL, although the network is trained by bag-level labels, the instance-level classifier can be trained. 
Let us consider a case using mean-pooling as the aggregation function; when the bag-level label is 0, the network is trained so that the output for an instance becomes 0, and when it's 1, positive instances are close to 1, while negative instances tend to keep 0 to make consistent with learning from the negative bags.

\noindent \textbf{LLP module with the positive instance mask. }
The LLP module aims to further classify instances, which are selected as positive by MIL, into $C$ classes.
To train the $C$ positive class classifier $h$, we use positive bags, their proportions for positive instances, and their instance scores estimated by MIL.
Since the given proportion is only for positive classes and the bag $\bB^i$ may contain negative instances, selecting only positive instances from the bag is required to compute the proportion loss.
As discussed above, to make the positive instance selection differentiable in joint learning and mitigate the adverse effect of MIL errors, we use a soft mask, which gives weight to an instance based on the positive score estimated by MIL, instead of a hard selection.

Specifically, the $C$-class classifier $h$ outputs the class probabilities $\hat{{\bf z}}^i_j = (\hat{z}_{j,1}^i,\ldots,$
$\hat{z}_{j,C}^i) \in \Delta^C$ of the instance $\bx_j^i$, where $\hat{z}_{j,c}^i=h(f(\bx_j^i))_c$ is the probability for class $c$ and $\sum_c{z_{j,c}^i}=1$. Then, the label proportions $\hat{\bp}^i \in \Delta^C$ of bag $\bB^i$ is estimated by the summation of multiplying the pos/neg instance score and the class probability as follows:
\begin{align}
\hat{\bp}^i = \frac{\sum_{j=1}^{|B^i|} \hat{s}_j^i \hat{{\bf z}}^i_j}{\sum_{j=1}^{|B^i|} \hat{s}_j^i }.
\end{align}
The estimated proportion can be considered as the proportion of the positive classes since the negative instances are ignored by multiplying the binary instance score, which is close to 0 if the MIL estimation is correct. This soft mask may mitigate the adverse effect for the LLP by false positives and negatives in the MIL estimation during training.

The proportion loss between the estimated proportion $\hat{\bp}^i$ and the ground truth $\bp^i$ of the positive classes is defined as:
\begin{align}
\mathcal{L}_\mathrm{LLP}(\bp^i, \hat{\bp}^i) = - \sum_{c=1}^C p_c^i \log \hat{p}_c^i.
\end{align}

\noindent \textbf{Simultaneous learning of MIL and LLP. }
All the modules $f$, $g$, $h$ are trained by the entire loss function $\mathcal{L}$ defined by the sum of the losses for LLP and MIL: 
$ \mathcal{L}_\mathrm{LLP}(\bp^i, \hat{\bp}^i) + w_\mathrm{MIL} \ \mathcal{L}_\mathrm{MIL}(Y^i, \hat{S}^i),$
where the hyperparameter $w_\mathrm{MIL}$ is the weight of $\mathcal{L}_\mathrm{MIL}$.

\noindent 
\textbf{Inference. }
In inference, an instance is classified into $C+1$ class by the two-stage approach using the trained networks.
First, the MIL module classifies an instance into a positive or negative instance. Then, when the instance is classified as positive, the instance is further classified into either of the $C$ positive classes.

\section{Experiments}

\noindent
\textbf{Comparative methods. }
We evaluated the following two methods as oracles of instance-level supervised learning (SL) and learning from ``complete'' label proportions: 1) cross-entropy loss (CE), which is a conventional classification method trained by a cross-entropy loss, assuming the class label is given for each instance as training data; and
2) proportion loss (PL), which is a conventional standard method for LLP that uses proportion loss, assuming the proportions for all classes are given as training data.
\par

The existing WSL methods are not feasible for comparing our method directly in the same problem setup as shown in Table~\ref{tab:related}. 
We thus prepared two baselines: 3) partial proportions loss (PPL), which is an extension of the proportion loss that calculates loss only for positive classes, i.e., it uses the estimation results of the $(C+1)$-class instance classifier trained by LLP and ignores the loss for the negative class; and 4) two-stage approach, which combines MIL and LLP. Unlike our method, the MIL is trained independently from LLP. 
These represent non-trivial extensions of existing methods and serve as strong baselines.
\par


\noindent
\textbf{Implementation details. }
All methods used ResNet 18 as the neural network model of each module.
The learning rate was set to $0.0003$, and the optimizer was Adam.
The batch size was set to $16$ bags. The model was trained until the validation loss did not update for $30$ epochs. 
The hyperparameter $w_\mathrm{MIL}$ was set to $0.01$. An optimal aggregation function in MIL was chosen from mean-pooling, max-pooling, and LSE-pooling using validation data. 
As performance metrics, we used instance-level accuracy~(Acc) and mean intersection over Union~(mIoU)~\cite{CsurkaG2013}.

\noindent \textbf{Dataset.}
To show the effectiveness of our method in pathological image segmentation, we evaluated our method using one public WSI dataset,  CRC100K~\cite{CRC100K}, and one private WSI dataset, Chemotherapy.
As preliminary experiments, we also evaluated our method using CIFAR10 and SVHN, which is detailed in the supplementary material.
\par

CRC100K contains $100,000$ 
image patches from WSIs with nine-class annotations.
In the experiment, we treated cancer-associated stroma and colorectal adenocarcinoma epithelium as two tumor classes and merged the remaining seven healthy classes as a non-tumor class because the proportions of tumor classes are only given in a real scenario. 
Each bag comprises $32$ samples whose label proportion was randomly determined. The $32$ instances were randomly extracted from the dataset according to the label proportion.
500 positive and 500 negative bags were prepared for training and validation. Other 100 positive and 100 negative bags were prepared for the test subset.\par
\par

In Chemotherapy, there are three classes: non-tumor (negative), tumor-bed (positive), and residual-tumor (positive). The viable tumor rate (the residual-tumor region size divided by the sum of the regions of tumor-bed and residual-tumor) is diagnosed in clinical. It indicates that the proportion of the non-tumor region is unknown, and the proportions of the positive classes are known as the partial proportion labels.
The chemotherapy dataset comprises $143$ positive WSIs and $10$ negative WSIs, which form bags. 
To show the effectiveness of our method, we prepared the segmentation masks for quantitative evaluation. Patients are separated in training, validation, and testing.
\par

\begin{table*}[t]
\caption{Accuracy and mIoU on pathological image datasets. \label{tab:quan_result_wsi}}
\centering
\begin{tabular}{lllcccc}
\hline
        &                                 &           & \multicolumn{2}{c}{CRC100K}                       & \multicolumn{2}{c}{Chemotherapy}                  \\ \cline{4-7} 
Setting & Given label                     & Method    & Acc.{[}\%{]} $\uparrow$ & mIoU{[}\%{]} $\uparrow$ & Acc.{[}\%{]} $\uparrow$ & mIoU{[}\%{]} $\uparrow$ \\ \hline
SL      & Instance label              & CE        & 99.22$\pm$0.12          & 97.98$\pm$0.34          & 86.31$\pm$2.10          & 70.62$\pm$3.44          \\
LLP     & (Comp.) label proportion   & PL        & 99.02$\pm$0.14          & 97.50$\pm$0.30          & 84.53$\pm$2.30          & 63.44$\pm$5.19          \\ \hline
LPLP    & Partial label proportion & PPL       & 96.29$\pm$0.65          & 90.82$\pm$1.50          & 72.08$\pm$6.20          & 48.01$\pm$8.98          \\
        &                                 & Two-stage & 83.96$\pm$3.61          & 61.34$\pm$8.11          & 50.49$\pm$6.85          & 27.26$\pm$3.16          \\
\rowcolor{Gray}        &                                 & Ours      & \textbf{98.53$\pm$0.39} & \textbf{96.32$\pm$0.96} & \textbf{74.10$\pm$3.07} & \textbf{52.01$\pm$3.77} \\ \hline
\end{tabular}
\end{table*}
\begin{figure*}[t]
\centering
\includegraphics[width=1.\linewidth]{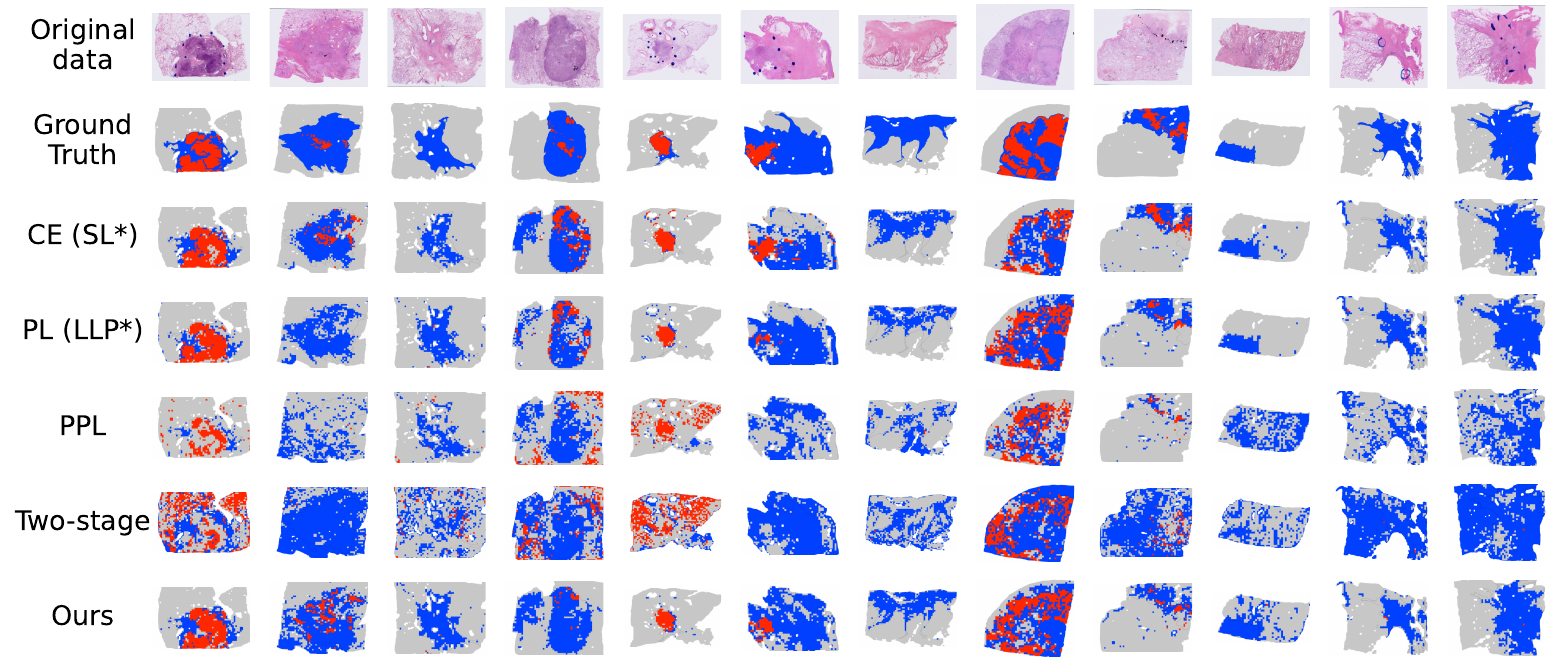}\\
\caption{Segmentation results.
\label{fig:wsi_segmentaion}
}
\end{figure*}

\noindent
\textbf{Quantitative evaluation results. }
Table~\ref{tab:quan_result_wsi} shows the accuracy and mIoU on the two pathological datasets: CRC100K and chemotherapy.
``Setting'' and ``Given label'' indicate the problem scenario and the type of labels used for training, respectively.
In CRC100K, despite the challenging LPLP scenario, our method achieved a competitive performance with the oracles (SL and LLP) results, which used more informative labels as training data.
In particular, the difference with LLP using ``complete'' label proportions was slight. Our method can successfully estimate the negative instances by MIL, and thus, the classification performance of positive classes was also achieved at the same level as LLP.
Next, our method was evaluated with two baseline methods in the same scenario (LPLP).
In both datasets, our method outperformed the baseline methods.
Our method was superior to PPL. We consider that the extension of the conventional proportion loss is insufficient for LPLP because it is difficult to train the discriminative features of negative instances from only the negative bags.
Moreover, compared to the two-stage method, 
it showed the effectiveness of our joint learning of MIL and LLP performed via differentiable positive-instance masks.
\par

\begin{table}[t]
\centering
\caption{Accuracy and mIoU in instance-level binary classification by the standard MIL  and the proposed method.\label{tab:MIL_classification}}
\begin{tabular}{lcccc}
\hline
       & \multicolumn{2}{c}{CRC100K}                       & \multicolumn{2}{c}{Chemotherapy}                  \\ \cline{2-5} 
 & Acc.{[}\%{]} $\uparrow$ & mIoU{[}\%{]} $\uparrow$ & Acc.{[}\%{]} $\uparrow$ & mIoU{[}\%{]} $\uparrow$ \\ \hline
MIL    & 85.14$\pm$3.78          & 68.88$\pm$7.35          & 47.09$\pm$7.05          & 26.24$\pm$7.22          \\
\rowcolor{Gray}Ours   & \textbf{97.98$\pm$1.13} & \textbf{95.57$\pm$2.43} & \textbf{69.44$\pm$6.58} & \textbf{51.16$\pm$5.96} \\ \hline
\end{tabular}
\end{table}

\noindent
\textbf{Effectiveness of joint learning for the performance of MIL. }
Table~\ref{tab:MIL_classification} shows the performances of binary classification by MIL in the two-stage method and our method.
Our method significantly improved both the Acc and mIoU in two datasets. It indicates that our simultaneous learning is synergistic, i.e., the loss from LLP has a good effect on the feature extractor $f$ and the binary classifier $g$ of MIL.

\noindent\textbf{Qualitative evaluation results. }
Fig.~\ref{fig:wsi_segmentaion} shows the segmentation results on Chemotherapy
: original image, ground truth, segmentation results from 1) CE, 2) PL, 3) PPL, 4) Two-stage, and ours from the top to the bottom row, respectively.
Gray indicates non-tumor areas, blue represents tumor-bed, and red signifies residual-tumor.
Our method's segmentation results were more accurate than the baseline methods in the LPLP scenario.
In addition, our method was competitive with SL and LLP, although a part of the label proportions was not given.

\section{Conclusion}
This paper first formulated the problem of learning from  {\em partial} label proportions (LPLP)  for the WSI segmentation task with the tumor subtype proportions. We then decomposed the LPLP problem into two WSL problems, MIL and LLP, and proposed their joint and end-to-end optimization method. The experimental results indicated that very weak but easily available supervision of the partial label proportions shows a similar performance to stronger supervision on the CRC100K dataset. Even on the chemotherapy dataset, our method outperformed other 
possible methods of LPLP. 


\bibliographystyle{splncs04}
\bibliography{main}


\end{document}


\begin{table*}[h]
\centering
\caption{
Experimental results on CIFAR10 and SVHN which have widely been used in previous papers on LLP.
Despite the challenging LPLP scenario, our method achieved a competitive performance with the oracles (SL and LLP) results, which used more informative labels as training data.
Our method outperformed the baseline methods (PPL and Two-stage) in the same scenario (LPLP). 
The performance of PPL was the worst. We consider that the extension of the conventional proportion loss is insufficient for LPLP.
The comparison with Two-stage shows the effectiveness of our joint learning of MIL and LLP.
\label{tab:quan_result_toy}}
\begin{tabular}{lllcccc}
\hline
        &                                 &           & \multicolumn{2}{c}{CIFAR10}                       & \multicolumn{2}{c}{SVHN}                          \\ \cline{4-7} 
Setting & Given label                     & Method    & Acc.{[}\%{]} $\uparrow$ & mIoU{[}\%{]} $\uparrow$ & Acc.{[}\%{]} $\uparrow$ & mIoU{[}\%{]} $\uparrow$ \\ \hline
SL      & Instance label $y$              & CE        & 80.44$\pm$0.84          & 53.31$\pm$1.32          & 88.91$\pm$2.03          & 70.42$\pm$4.24          \\
LLP     & (Comp.) label proportion   & PL        & 78.78$\pm$1.30          & 49.04$\pm$3.17          & 89.32$\pm$1.27          & 69.48$\pm$2.97          \\ \hline
LPLP    & Partial label proportion  & PPL       & 70.72$\pm$0.91          & 34.44$\pm$1.58          & 80.66$\pm$4.37          & 53.51$\pm$9.23          \\
        &                                 & Two-stage & 76.76$\pm$1.51          & 46.54$\pm$2.71          & 85.25$\pm$2.94          & 61.40$\pm$6.66          \\
\rowcolor{Gray}        &                                 & Ours      & \textbf{77.04$\pm$0.64} & \textbf{49.18$\pm$1.02} & \textbf{88.05$\pm$0.83} & \textbf{68.82$\pm$2.02} \\ \hline
\end{tabular}
\end{table*}

\begin{figure}[h]
\centering
\includegraphics[width=.6\linewidth]{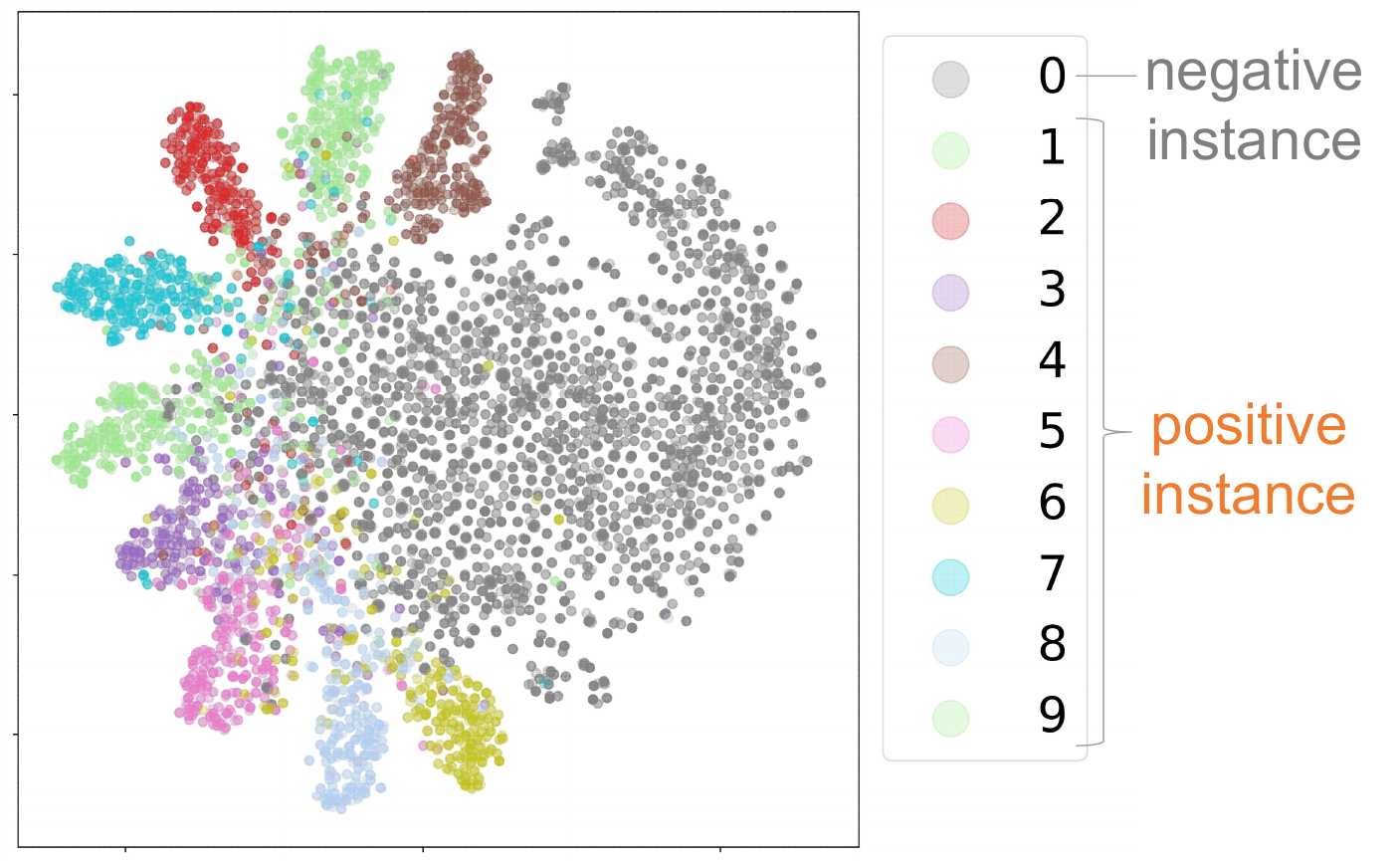}
\caption{
Feature distributions of the test samples from SVHN by t-SNE. Each color indicates each class, where gray shows the negative samples.
The negative instances (gray) were successively separated from the positive instances (non-gray). This indicates that our method successfully trained the MIL classifier.
Furthermore, the feature distributions of sub-classes of the positive class are separated successfully. It indicates the LLP module after MIL works well.
}
\label{fig:vis_tsne}
\end{figure}

\begin{figure}[h]
\centering
\subfigure[CIFAR10]{%
\includegraphics[width=.4\linewidth]{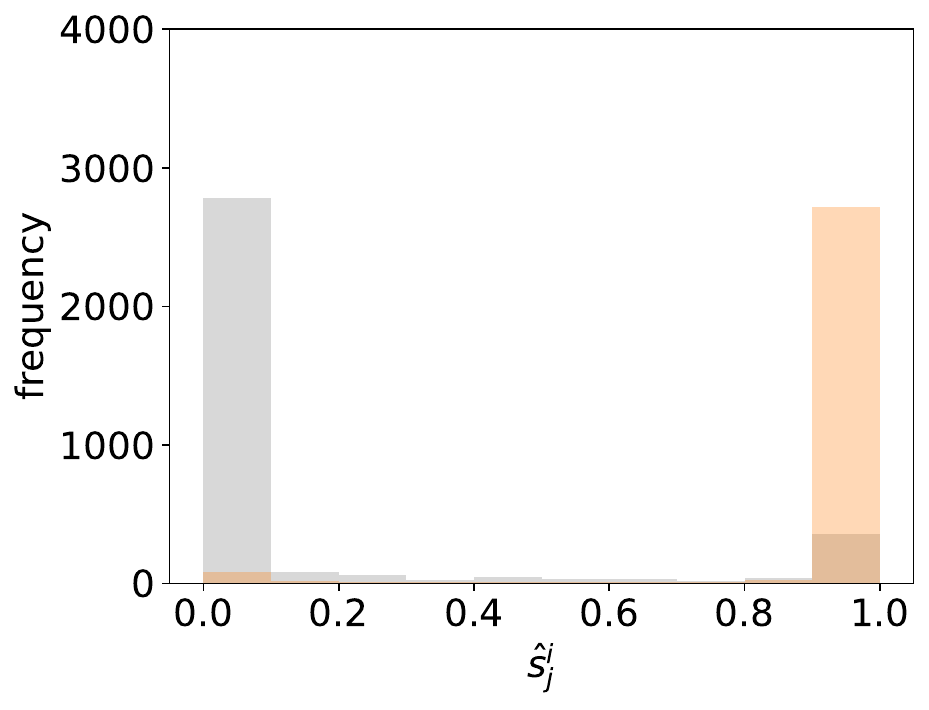}}%
\subfigure[SVHN]{%
\includegraphics[width=.4\linewidth]{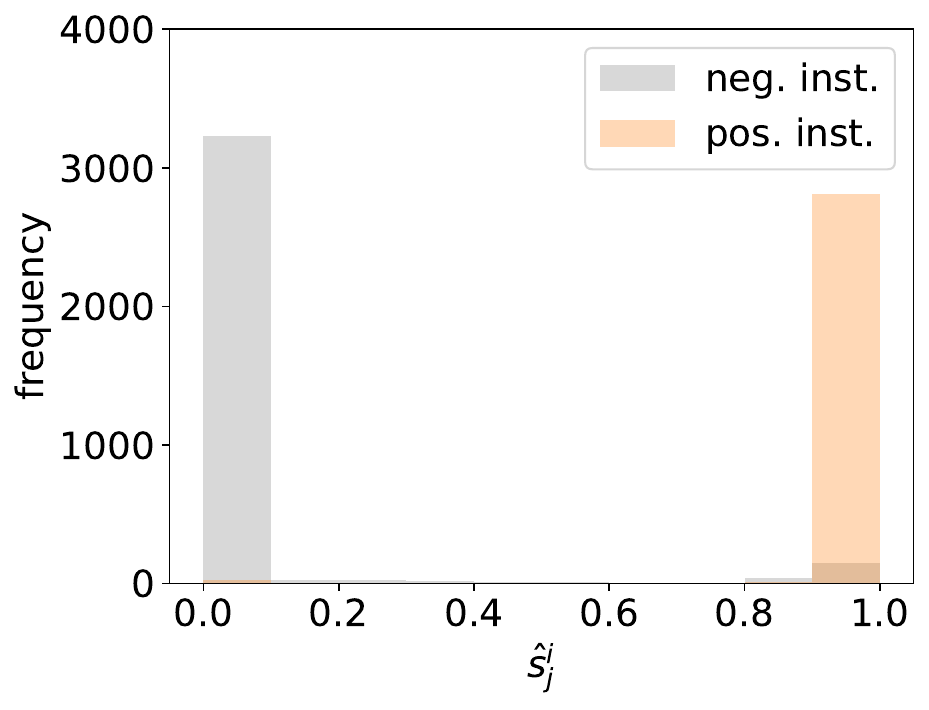}}%
\caption{
Histogram of the estimated positive score $\hat{s}_i^j$ by MIL in CIFAR10 and SVHN.
In the results, almost all instances are successfully classified.
In addition, the estimated score tends to take 0 or 1 after training, although the score can take a value from 0 to 1 $\hat{s}^i_j \in [0,1]$ to give weight for each instance during training.
It indicates that the soft mask can work as a hard mask in the end of training.
}
\label{fig:histo_mask}
\end{figure}

\begin{table*}[h]
\caption{
Experimental results when changing the bag size (32, 64, 128) while the number of bags was fixed.
Our method was the best in all bag sizes.
It shows the robustness of our method for the bag size.
\label{tab:bag_size}}
\centering
\begin{tabular}{llcccccc}
\hline
\multicolumn{2}{l}{Bag size} & \multicolumn{2}{c}{32}                            & \multicolumn{2}{c}{64}                            & \multicolumn{2}{c}{128}                           \\ \hline
Setting      & Method        & Acc.{[}\%{]} $\uparrow$ & mIoU{[}\%{]} $\uparrow$ & Acc.{[}\%{]} $\uparrow$ & mIoU{[}\%{]} $\uparrow$ & Acc.{[}\%{]} $\uparrow$ & mIoU{[}\%{]} $\uparrow$ \\ \hline
SL           & CE            & 80.44$\pm$0.84          & 53.31$\pm$1.32          & 80.29$\pm$0.51          & 54.92$\pm$1.16          & 81.17$\pm$0.64          & 57.69$\pm$0.87          \\
LLP          & PL            & 78.78$\pm$1.30          & 49.04$\pm$3.17          & 79.39$\pm$0.97          & 50.23$\pm$1.60          & 81.08$\pm$0.73          & 53.61$\pm$1.72          \\ \hline
LPLP         & PPL           & 70.72$\pm$0.91          & 34.44$\pm$1.58          & 74.12$\pm$0.83          & 41.56$\pm$1.38          & 75.21$\pm$1.10          & 44.36$\pm$2.05          \\
             & Two-stage     & 76.76$\pm$1.51          & 46.54$\pm$2.71          & 78.19$\pm$1.05          & 49.34$\pm$1.86          & 79.00$\pm$0.81          & 52.56$\pm$1.20          \\
  \rowcolor{Gray}           & Ours          & \textbf{77.04$\pm$0.64} & \textbf{49.18$\pm$1.02} & \textbf{78.95$\pm$1.21} & \textbf{52.45$\pm$1.29} & \textbf{80.47$\pm$0.72} & \textbf{55.13$\pm$0.76} \\ \hline
\end{tabular}
\end{table*}

\begin{table}[h]
\caption{
Experimental results when using a different LLP backbone, LLP-VAT, instead of PL on CIFAR10 to show the extension ability of our method.
In this experiment, LLP-VAT is an oracle that uses the complete label proportion. The backbone method for LLP was changed to LLP-VAT in two baseline methods and our method. Our method outperformed the comparative baseline methods and achieved comparative performances with the oracle. Any LLP methods can be applied to our method.}
\label{tab:extention_llp}
\centering
\begin{tabular}{lcc}
\hline
Method                      & Acc.{[}\%{]} $\uparrow$                     & mIoU{[}\%{]} $\uparrow$                     \\ \hline
LLP-VAT                     & 78.81$\pm$1.53          & 48.73$\pm$3.24          \\ \hline
PPL (w/ LLP-VAT)       & 71.08$\pm$0.72          & 34.92$\pm$1.56          \\
         Two-stage (w/ LLP-VAT) & 77.39$\pm$1.02          & 46.50$\pm$2.06          \\
\rowcolor{Gray}         Ours (w/ LLP-VAT)      & \textbf{77.68$\pm$0.87} & \textbf{49.21$\pm$0.78} \\ \hline
\end{tabular}
\end{table}


%% file: preamble.tex
\usepackage[dvipsnames]{xcolor}

\usepackage{times}
\usepackage{epsfig}
\usepackage{graphicx}
\usepackage{amsmath}
\usepackage{amssymb}
\usepackage{subfigure}
\usepackage{booktabs}
\usepackage{color}
\usepackage{comment}
\usepackage{colortbl}
\usepackage{array}
\usepackage{xcolor}
\definecolor{Gray}{gray}{0.9}
\usepackage{arydshln}
\usepackage{multirow}
\usepackage{here}
\usepackage[normalem]{ulem}
\usepackage{bm} 
\useunder{\uline}{\ul}{}

\newcommand{\bx}{{\bm x}}

\newcommand{\bB}{{\bm B}}

\newcommand{\bp}{{\bm p}}

\renewcommand{\paragraph}[1]{\noindent\textbf{#1}\quad}